\pdfoutput=1

\documentclass[11pt]{article}

\usepackage[]{acl}

\usepackage{times}
\usepackage{latexsym}

\usepackage[T1]{fontenc}

\usepackage[utf8]{inputenc}

\usepackage{microtype}

\usepackage{inconsolata}

\usepackage{caption}
\usepackage{subcaption}
\usepackage{enumitem}
\usepackage{flushend}
\usepackage{balance}
\usepackage{lineno}
\usepackage{amsmath,amssymb,amsfonts}
\usepackage{algorithm}
\usepackage{algorithmic}
\usepackage{graphicx}
\usepackage{textcomp}
\usepackage{xcolor}
\usepackage{colortbl}
\usepackage{amsthm}
\usepackage{url}
\usepackage{float}
\usepackage{multirow}
\usepackage{multicol}
\usepackage{color}
\usepackage{bm}
\usepackage{bbm}

\usepackage{pifont}
\usepackage{array}
\newcolumntype{L}[1]{>{\raggedright\let\newline\\\arraybackslash\hspace{0pt}}m{#1}}
\newcolumntype{C}[1]{>{\centering\let\newline  \\\arraybackslash\hspace{0pt}}m{#1}}
\newcolumntype{R}[1]{>{\raggedleft\let\newline \\\arraybackslash\hspace{0pt}}m{#1}}

\usepackage{fontawesome}
\usepackage{arydshln}

\definecolor{color1}{rgb}{0.1,0.7,0.8} 
\definecolor{color2}{rgb}{0.9,0.1,0.1} 
\definecolor{color3}{rgb}{0.7,0.3,0.7} 
\definecolor{color4}{rgb}{0.3,0.3,0.7} 
\definecolor{color5}{RGB}{8, 102, 3} 
\definecolor{color6}{rgb}{0.53, 0.66, 0.42} 

\newcommand\llamaex{LLaMA-DLO}

\usepackage[utf8]{inputenc} 
\usepackage[T1]{fontenc}    
\usepackage{hyperref}       %
\hypersetup{colorlinks,linkcolor={red}}        %
\usepackage{url}            
\usepackage{booktabs}       
\usepackage{amsfonts}       
\usepackage{nicefrac}       
\usepackage{microtype}      
\usepackage{xcolor}         
\usepackage{comment} 
\usepackage{mdframed}
\usepackage{tikz}

\title{DLO: Dynamic Layer Operation for Efficient Vertical Scaling of LLMs}

\author{
  Zhen Tan\textsuperscript{\ding{171}}\thanks{\ \ Equal contribution.} \quad 
  Daize Dong\textsuperscript{\ding{170}}\footnotemark[1] \quad 
  Xinyu Zhao\textsuperscript{\ding{169}} \quad 
  Jie Peng\textsuperscript{\ding{167}} \quad 
  \textbf{Yu  Cheng}\textsuperscript{\ding{168}} \quad
  \textbf{Tianlong Chen}\textsuperscript{\ding{169}} \\
  \textsuperscript{\ding{171}}School of Computing, and Augmented Intelligence, Arizona State University\\
  \textsuperscript{\ding{170}}Shanghai Artificial Intelligence Laboratory \\
  \textsuperscript{\ding{168}}Department of Computer Science, Chinese University of Hong Kong \\
  \textsuperscript{\ding{167}}School of Artificial Intelligence and Data Science, University of Science and Technology of China \\
  \textsuperscript{\ding{169}}Department of Computer Science, University of North Carolina at Chapel Hill\\
  {\tt ztan36@asu.edu} \quad {\tt dongdaize.d@pjlab.org.cn} \quad {\tt  chengyu@cse.cuhk.edu.hk}\\
  {\tt \{xinyu,tianlong\}@cs.unc.edu} \quad
  {\tt pengjieb@mail.ustc.edu.cn} \quad
  \\
}

\newmdenv[
    linewidth=2pt,       
    roundcorner=10pt,    
    linecolor=black,     
    backgroundcolor=gray!10, 
    skipabove=5pt,      
    skipbelow=5pt       
]{custombox}

\begin{document}

\maketitle

\begin{abstract}
In this paper, we introduce Dynamic Layer Operations (DLO), a novel approach for vertically scaling transformer-based Large Language Models (LLMs) by dynamically expanding, activating, or skipping layers using a sophisticated routing policy based on layerwise feature similarity.  Unlike traditional Mixture-of-Experts (MoE) methods that focus on extending the model width, our approach targets model depth, addressing the redundancy observed across layer representations for various input samples. Our framework is integrated with the Supervised Fine-Tuning (SFT) stage, eliminating the need for resource-intensive Continual Pre-Training (CPT). Experimental results demonstrate that DLO not only outperforms the original unscaled models but also achieves comparable results to densely expanded models with significantly improved efficiency. Our work offers a promising direction for building efficient yet powerful LLMs. We will release our implementation and model weights upon acceptance.
\end{abstract}

\section{Introduction}
\begin{figure}
    \centering
    \scalebox{0.96}{
\includegraphics[width=\linewidth]{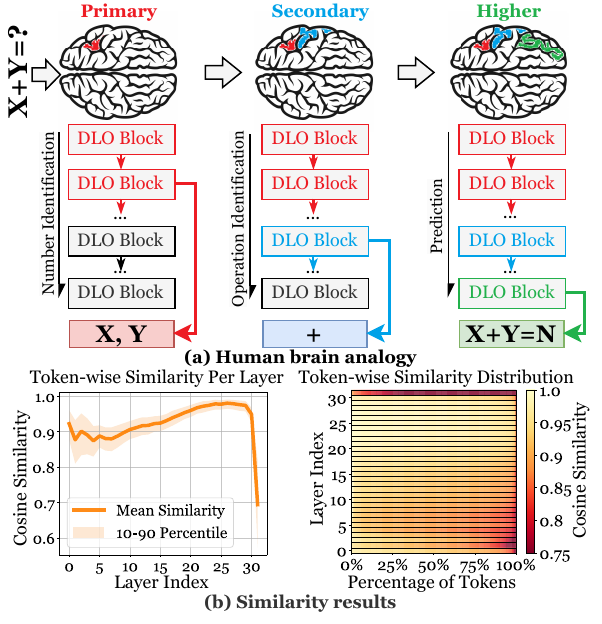}}
\vspace{-4mm}
    \caption{\small (a) DLO structure that ensembles human brain activities in a math problem example~\cite{koechlin2003architecture}, where the primary neurons preceive numbers, secondary neurons understand operations, and high-order neurons calclulate the results. (b)~Layer-Wise token similarity and distribution.}
    \label{fig:teaser}
    \vspace{-7mm}
\end{figure}
Large Language Models (LLMs)~\cite{achiam2023gpt,team2023gemini} have shown remarkable success across various natural language processing (NLP) tasks~\cite{hadi2023survey,tan2024large,li2024facial,li2024dalk}, leveraging their vast capacity to capture complex patterns in data. Traditional scaling of these models has predominantly focused on \textit{horizontal} expansion, as seen in Mixture-of-Experts (MoE) architectures~\cite{shazeer2017outrageously,fedus2022switch,lepikhin2020gshard}, where the width of the model is increased by adding more experts. 
This approach primarily optimizes parameter usage and computational cost by activating a fixed portion of parameters conditioned on the given input~\cite{fedus2022review}. 

However, the potential for \textit{vertical} expansion remains underexplored. Inspired by how the human brain allocates more neurons for complex tasks and forms deeper neural chains~\cite{baddeley1992working,koechlin2003architecture}, we propose focusing on vertical scaling. Our method dynamically expands, activates, or skips layers to optimize model depth and reduce redundancy, as shown in Figure~\ref{fig:teaser}~(a).


There are three critical challenges in vertically scaling LLMs: \ding{182}~\textbf{Optimization Complexity.} Dynamically adding or pruning layers  making the process hard to optimize. Obtaining optimal such operations have been proved to be a NP-hard problem~\cite{glorot2010understanding,hestness2017deep}, while an approximation method~\cite{wang2023learning} has shown compromised improvement.
\ding{183}~\textbf{Computation Cost.}
The inherent computational cost is associated with processing deeper networks. Each additional layer contributes to the overall latency and resource consumption.
\ding{184}~\textbf{Feature Collapse.} Our analysis in Figure~\ref{fig:teaser}~(b) reveals that for a significant number of inputs, the representations across consecutive layers exhibit substantial similarity, suggesting that many layers may be redundant for certain samples.

To address these challenges, in this paper, we propose \underline{\textbf{D}}ynamic \underline{\textbf{L}}ayer \underline{\textbf{O}}peration (\textbf{DLO}), that consists of three operations: (\textbf{\textit{i}})~\texttt{expansion}, (\textbf{\textit{ii}})~\texttt{activation}, and (\textbf{\textit{iii}})~\textit{\texttt{skipping}}, for dynamic vertical scaling of LLMs without a proportional increase in computational cost.
Our specific designs are as follows: \ding{182}~\texttt{\textbf{Expansion}}: \textit{Additional layers} are dynamically expanded from existing ones, easing optimization complexity. 
\ding{183}~\textbf{\texttt{Activation} \& \texttt{Skip}}: \textit{Feature Similarity} guides the activation and skipping of layers. We propose similarity-induced labels to train the router that controls these operations.
\ding{184}~\textbf{Adaptive {FLOPs}}: \textit{Sparsity settings} vary for layers facilitate adaptive FLOPs for different tokens, maintaining efficiency.
\ding{185}~\textbf{Enhanced Generalizability}:\textit{ Layer-specific learning rates}, based on sparsity, further improve the model's ability to generalize across tasks.
\textbf{\textit{Note}} that all modules are trained during the Supervised Fine-Tuning (SFT) stage, eliminating the need for Continual Pre-training (CPT) and simplifying the training process.
Our primary contributions are as follows:
\begin{itemize}[leftmargin=*,itemsep=1pt]
    \vspace{-2mm}
    \item \underline{\textbf{\textit{Method.}}} We introduce a novel method, \textbf{DLO}, for dynamically scaling LLMs vertically by dynamically expanding, activating, or skipping layers.
    \vspace{-6mm}
    \item \underline{\textbf{\textit{Performance \& Efficiency.}}} Through rigoerous experiments, we demonstrate that \textbf{DLO} not only surpasses the performance of the original unscaled models but also achieves comparable results to densely expanded models with significantly enhanced efficiency.
    \vspace{-2mm}
    \item \underline{\textbf{\textit{Applicability.}}} Fine-tuned on \textit{language understanding}, \textit{math}, and \textit{coding} tasks, we manifest DLO's effectiveness across multiple NLP tasks.
\end{itemize}

\section{Related Work}
\subsection{Mixture-of-Experts (MoE)} 
MoE architectures have emerged as a promising approach for enhancing the efficiency and scalability of LLMs~\cite{shazeer2017outrageously}. Traditional neural networks activate all parameters for every input, leading to significant computational overhead, particularly as models scale up. In contrast, MoE models activate only a subset of parameters for each input, optimizing computational resource usage and enabling models to scale to billions of parameters without a corresponding increase in computational cost per input~\cite{lepikhin2020gshard,fedus2022switch,zoph2022st,llama-moe-2023,jiang2024mixtral，zhu2024llama}.
This selective activation makes MoE highly efficient for both training and inference by focusing on horizontal expansion and adding more experts. However, MoE’s primary aim is to optimize width, potentially leaving layer redundancy unaddressed. Our Dynamic Layer Operation (DLO) approach complements MoE by focusing on vertical scaling through dynamic layer expansion and activation, targeting depth scalability and reducing potential feature redundancy. 



\vspace{-1mm}
\subsection{Efficient Model Stacking} 
Model stacking is a common ensemble learning technique that improves predictive performance by combining multiple models to leverage their complementary strengths~\cite{ting1997stacking, chen2015net2net}. In the context of LLMs, stacking can involve integrating various models into a hierarchical structure, where outputs from one model serve as inputs to another, capturing a broader range of features and patterns~\cite{dabre2019recurrent, chen2021bert2bert, wang2023lemon, kim2023solar}.

Recent advancements have focused on progressively stacking pre-trained transformer or self-attention layers to create composite language models~\cite{gong2019efficient, gu2020transformer, shen2022staged, evci2022gradmax, yao2023masked, du2024stacking, wu2024llama}. This approach reduce training costs by reusing pre-trained components. However, the increased depth and complexity of stacked models lead to high inference latency.

To mitigate this issue, layer-skipping methods have been developed, allowing models to ``early exit'' using additional layer-wise classifiers, thereby reducing the number of layers processed during inference~\cite{wang2022skipbert, chen2023ee, zhangconditional}. More recently, conditional computation techniques have been proposed to dynamically skip layers based on token-specific conditions, further enhancing efficiency~\cite{ainslie2023colt5, raposo2024mixture}. However, these methods often require modifications during the pre-training stage, adding computation complexity and limiting their application to existing pre-trained LLMs.
In contrast, our DLO method focuses on efficiency and scalability through dynamic vertical scaling within a single model during the SFT stage. 
It provides a comprehensive, high-performance solution to scaling LLMs without the extensive computational demands associated with stacked ensembles.



\section{Methodology}

\begin{figure}
    \centering
    \includegraphics[width=\linewidth]{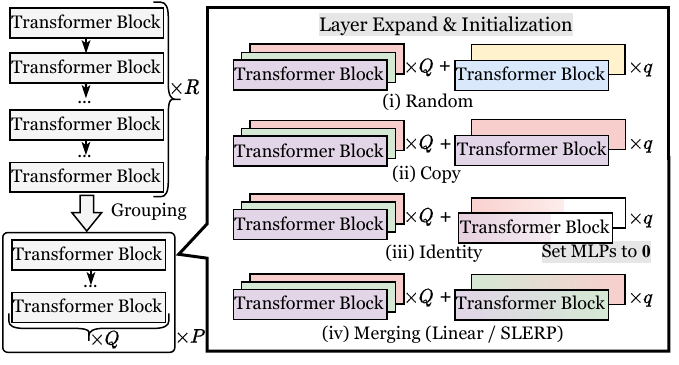}
    \caption{\small Layer extension with initialization strategies.}
    \label{fig:layer_expansion}
    \vspace{-4mm}
\end{figure}

In this section, we introduce the \underline{\textbf{D}}ynamic \underline{\textbf{L}}ayer \underline{\textbf{O}}peration (\textbf{DLO}) framework for efficienct vertical scaling of LLMs. DLO consists of three key operations: \texttt{expansion}, \texttt{activation}, and \texttt{skipping}. These operations dynamically adjust the model structure during the Supervised Fine-Tuning (SFT) phase to optimize computational efficiency and improve performance. A pseudo code style description is included in Appendix~\ref{app:code}.

\subsection{Layer Expansion}\label{sec:expand}

To facilitate dynamic depth adjustment, we introduce a group-based layer expansion strategy. Suppose the LLM has $R$ transformer layers, which we group into $P$ groups with $Q$ layers each, such that $R = P \times Q$. Each group is expanded to $Q' = Q + q$ layers, where $q$ is the number of additional layers introduced per group. The resulting number of layers will be $R' = P \times Q'$.

Let $\mathcal{G}_i$ denote the $i$-th group with layers $\mathcal{L}_{i1}, \mathcal{L}_{i2}, \ldots, \mathcal{L}_{iQ}$. The expanded group $\mathcal{G}_i'$ will contain layers $\mathcal{L}_{i1}, \mathcal{L}_{i2}, \ldots, \mathcal{L}_{i(Q+q)}$.
The expanded layers are initialized using a policy $\Pi$, and we consider several initialization strategies:
\begin{itemize}[leftmargin=*,itemsep=1pt]
    \vspace{-1mm}
    \item \textbf{Random Initialization ($\Pi_\text{rand}$)}: Initialize the new layers' weights $\theta'_{ij}$ using Xavier initialization~\cite{glorot2010understanding}.
    \begin{equation*}
        \theta'_{ij} \sim \mathcal{U}\left(-\sqrt{\frac{6}{n_{in} + n_{out}}}, \sqrt{\frac{6}{n_{in} + n_{out}}}\right), 
    \end{equation*}
    where $\forall j \in \{Q+1, Q+2, \ldots, Q+q\}$,     $\mathcal{U}$ denotes the uniform distribution, $n_{in}$ is the number of input units, and $n_{out}$ is the number of output units in the layer.

    \vspace{-2mm}
    \item \textbf{Copy from Previous Layer ($\Pi_\text{copy}$)}: Copy the parameters from the preceding layer.
    \begin{equation*}
        \theta'_{ij} = \theta_{i(Q+q-1)},  \forall j \in \{Q+1, \ldots, Q+q\}.
    \end{equation*}
    \vspace{-2mm}
     \item \textbf{Identity Initialization ($\Pi_\text{Identity}$)}~\cite{wu2024llama}: Copy from the preceding layer but set the output linear matrix of the multi-head self-attention (MHSA) to zero.
    \begin{enumerate}[leftmargin=*,itemsep=1pt,label=\alph*.]
    \vspace{-2mm}
        \item Copy the parameters of the previous layer:
        \begin{equation*}
            \theta'_{ij} = \theta_{i(Q+q-1)},  \forall j \in \{Q+1, \ldots, Q+q\}.
            \vspace{-2mm}
        \end{equation*}
        \item Set the weights of the output linear layer $W'_{\text{out}}$ in the MHSA to zero: $
            W'_{\text{out}} = 0.$
    \vspace{-1mm}
    \end{enumerate}
    In this way, the output of the expanded layers will preserve the features from the original layers.

    \vspace{-1mm}
    \item \textbf{Linear Merge ($\Pi_\text{linear}$)}: Merge from the preceding $\tau$ layers using a linear function.
    \vspace{-1mm}
    \begin{equation*}
        \theta'_{ij} = \sum_{k=1}^{\tau} \alpha_k \theta_{i(Q+q-k)},  \sum_{k=1}^{\tau} \alpha_k = 1.
    \vspace{-2mm}
    \end{equation*}

    \item \textbf{Spherical Linear Interpolation (SLERP) ($\Pi_\text{slerp}$)}~\cite{shoemake1985quaternion}: Merge from the preceding $\tau$ layers using SLERP. The SLERP method smoothly interpolates between two weight vectors on a unit sphere, maintaining constant velocity.
    The interpolation between two weight vectors $\mathbf{u}$ and $\mathbf{v}$ is defined as:
    \begin{equation*}
        \text{SLERP}(\mathbf{u}, \mathbf{v}, \alpha) = \frac{\sin((1 - \alpha) \Omega)}{\sin(\Omega)} \mathbf{u} + \frac{\sin(\alpha \Omega)}{\sin(\Omega)} \mathbf{v},
    \end{equation*}
    where $\Omega$ is the angle between $\mathbf{u}$ and $\mathbf{v}$:
    \begin{equation*}
        \Omega = \arccos \left( \frac{\mathbf{u} \cdot \mathbf{v}}{\|\mathbf{u}\| \|\mathbf{v}\|} \right),
    \end{equation*}
    and $\alpha \in [0, 1]$ is the interpolation parameter.

    In our context, for the new layer $j$, the parameters are initialized by interpolating between the weights of the previous layers $\theta_{i(Q+q-1)}$ and $\theta_{i(Q+q-\tau)}$:
    \begin{equation*}
        \theta'_{ij} = \text{SLERP}(\theta_{i(Q+q-1)}, \theta_{i(Q+q-\tau)}, \alpha),
    \end{equation*}
    where $\alpha$ controls the interpolation. 
    This ensures a smooth transition between layers, aiding in gradient flow and stable training. 
\end{itemize}
\vspace{-1mm}
We conduct comprehensive experiments on the choice of the policy $\Pi$ in Section~\ref{sec:ablation}.

\subsection{Layer Activation \& Skipping}

DLO dynamically skips the multi-layer perceptron (MLP) module within the transformer layer $\mathcal{L}_i$ for input tokens.
To achieve this, we uses a linear router to determine the activation of layers. 
Suppose we have the set of token embeddings in a sequence of length $S$ for a given layer $\mathcal{L}_i$, that is $\mathbf{h}_i = \{\mathbf{h}^s_i | s\in \mathbb{N}^*, s \leq S \}$, where in following contents we omit the superscript $s$ for better readability.
Considering feature redundancy, we use the router weights $W_i$ to process the input token $\mathbf{h}_i$ and obtain the decision score $r_i$, which is given by:
\begin{equation}\label{eq:router}
    r_i = \frac{\beta+(2\sigma(\mathbf{h}_i W_i) - 1)\gamma}{2} \in \Big(\frac{\beta-\gamma}{2}, \frac{\beta+\gamma}{2}\Big),
\end{equation}
where $\sigma$ is the sigmoid function, $\beta$ and $\gamma$ are hyperparameters controlling the output range. 
During the inference stage, this score $r_i$ determines whether layer $\mathcal{L}_i$ is active: the layer is activated if and only if $\mathcal{L}_i \geq \frac{\beta}{2}$, otherwise it's skipped. The final activated output for the layer is:
\begin{equation}\label{eq:output_calculation}
    \mathbf{h}_{i+1} = 
    \begin{cases}
        r_i \cdot \mathcal{M}_i \circ \mathcal{A}_i(\mathbf{h}_i) & \text{if } r_i \geq \frac{\beta}{2}, \\
        \mathcal{A}_i(\mathbf{h}_i) & \text{otherwise},
    \end{cases}
\end{equation}
where $\mathcal{M}_i$, $\mathcal{A}_i$ are the MLP and the attention modules within layer $\mathcal{L}_i$, respectively.
This activation \& skipping mechanism aims to encourage the utilization of the most relevant layers, thus reducing unnecessary computation.
In this paper we initialize the router weights $W_i$ as zeros and set $\beta=2.0,\gamma=0.05$, so that $r_i = 1.0$ on the first step and $r_i \in (0.975, 1.025)$ during training. This ensures benign initialization for activated tokens and avoids excessive disturbance on activated outputs brought by the decision scores.

\subsection{Training and Integration}
\label{sec:dlo_training}

\begin{figure}
    \centering
    \includegraphics[width=\linewidth]{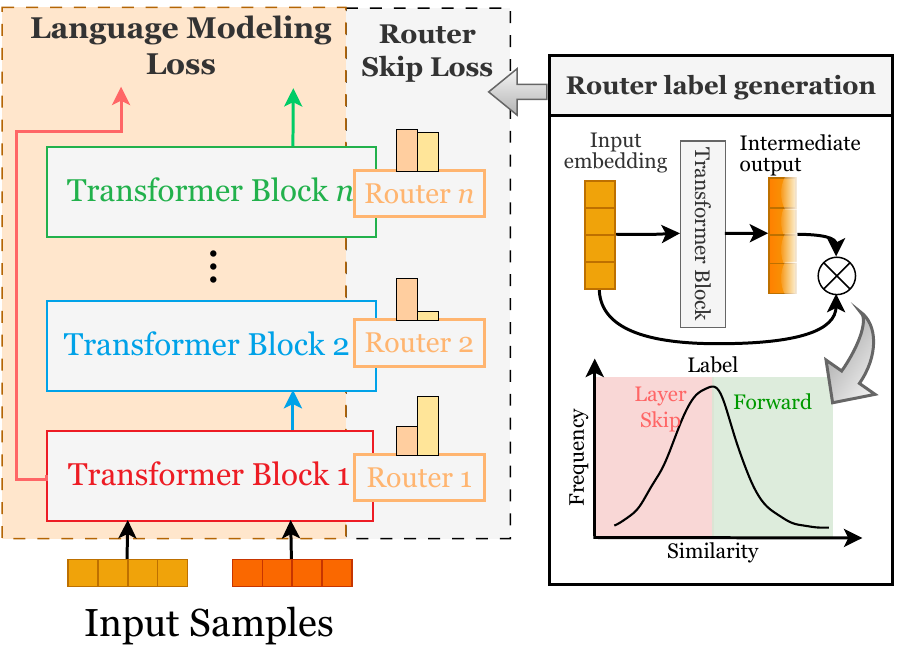}
    \caption{\small Training pipeline of \textbf{DLO}, consists of the downstream task loss and an auxilliary router skip loss supervised by generated router labels.}
    \label{fig:training}
    \vspace{-4mm}
\end{figure}

\paragraph{Similarity-induced Label \& Router Skip Loss.}
Given a pre-defined overall spasity $\rho$, we define a sparsity factor $\rho_i$ for each layer $\mathcal{L}_i$ that controls the layer-wise actived tokens. To determine the status of token $\mathbf{h}_i^s$, we utilize predicted router labels $\hat\Lambda^s = \{\hat\lambda_{1}^s, \dots, \hat\lambda_{i}^s, \dots, \hat\lambda_{R'}^s\}$, which is obtained through the decision scores $r_i^s$:
\begin{equation}\label{eq:gate}
    \hat\lambda_i^s = 
    \begin{cases}
        1 & \text{if } r_i^s \in \mathrm{Top}_{\lfloor(1-\rho_i) S\rfloor}\big(\{r_i^s\}_{s=1}^S\big), \\
        0 & \text{otherwise}.
    \end{cases}
\end{equation}
where $\hat\lambda_i^s=1$ indicates layer $\mathcal{L}_i$ is predicted to be activated, and vice versa. 
To train the routers, we utilize the supervised router labels $\Tilde\Lambda^s = \{\Tilde\lambda_{1}^s, \dots, \Tilde\lambda_{i}^s, \dots, \Tilde\lambda_{R'}^s\}$ to guide the learning of sparsity, \textit{i.e.}, to skip or not. The router labels $\Tilde\lambda_{i}^s$ of layer $\mathcal{L}_i$ at training step $t$ is determined by the following procedures:
\begin{enumerate}[leftmargin=*,itemsep=1pt]
    \item The cosine feature similarity of layer $i$ is calculated through features across MLP $\mathcal{M}_i$:
    \begin{equation}
        \mu_{i}^s = \frac{\mathcal{A}_i(\mathbf{h}_i^s) \cdot \mathcal{M}_i \circ \mathcal{A}_i(\mathbf{h}_i^s)}{\|\mathcal{A}_i(\mathbf{h}_i^s)\| \|\mathcal{M}_i \circ \mathcal{A}_i(\mathbf{h}_i^s)\|} \in [0, 1].
    \end{equation}
    \item The similarity are sorted over all the layers, and the similarity-induced label is given as follows:
    \begin{equation}
    \small
        \Tilde\lambda_{i}^s = 
        \begin{cases}
            1 & \text{if } \mu_{i}^s \in \mathrm{Bottom}_{\lfloor(1-\rho)R'S\rfloor}\big(\{\mu_{i}^s\}_{i,s=1}^{R',S}\big), \\
            0 & \text{otherwise.}
        \end{cases}
    \end{equation}
    where $\mathrm{Bottom}_{\lfloor(1-\rho)R'S\rfloor}$ indicates the labels for tokens with the least $\lfloor(1-\rho)R'S\rfloor$ portion of cosine similarity are set to $1$s, which are expected to be activated. 
    \item A skip loss $\mathcal{L}_\text{skip}$ based on the Binary-Cross-Entropy loss $\mathcal{L}_{\text{BCE}}$ is incorporated to guide the learning of the router attached to each layer:
    \begin{equation}
        \mathcal{L}_{\text{skip}} = \frac{1}{R'S} \sum_{i,s=1}^{R',S}\mathcal{L}_{\text{BCE}}(\sigma(\mathbf{h}_i^sW_i), \Tilde\lambda_{i}^s).
    \end{equation}
\end{enumerate}
Given the task-specific loss $\mathcal{L}_{\text{task}}$, the overall loss function for DLO training is:
\begin{equation}
    \mathcal{L} = \mathcal{L}_{\text{task}} + \mathcal{L}_{\text{skip}}. 
\end{equation}

\paragraph{Skip Rate Dynamics.}
The redundancy exhibits an imbalanced distribution across layers, as is evidenced in Figure~\ref{fig:teaser} (b). To this end, we adjust the next-step $\rho_{i,t+1}$ for each training step $t$ over the total $T$ steps, where the initial skip rate $\rho_{i,1}=\rho$. The layer-wise sparsity factor $\rho_{i,t+1}$ is calculated using the router labels as follows:
\begin{equation}
    \rho_{i,t+1} = \frac{\sum_{s=1}^S \Tilde\lambda_{i,t}^s}{S},
\end{equation}
where $\Tilde\lambda_{i,t}^s$ is the supervised router label of the $s$-th token in layer $\mathcal{L}_i$ at step $t$.
Additionally, we employ an annealing technique on the skip rate to ensure the warm start. During training, the overall skipping rate gradually increases from an initial low value $\bar{\rho}$ to the target sparsity level $\rho$ over a predefined number of steps $T'$. The overall skip rate $\rho^{t}$ at step $t$ is given by:
\begin{equation}
    \rho^{t} =
    \begin{cases}
        \bar{\rho} + \left( \rho - \bar{\rho} \right) \frac{t}{T'} & \text{if } t \leq T', \\
        \rho & \text{otherwise},
    \end{cases}
\end{equation}
where we set $\bar{\rho} = 0$. This annealing process helps the model to progressively adapt to higher sparsity levels with smoother changes, leading to more stable training and better convergence. 

\paragraph{Layer-Wise Learning Rates.}
DLO also employs layer-wise learning rates $\zeta_{i,t}$, adjusted based on sparsity to promote generalizability. The learning rate for each layer is defined as:
\begin{equation}
    \zeta_{i,t} = \bar\zeta \cdot \frac{1 - \rho_{i,t}}{1 - \rho^{t}},
\end{equation}
where $\bar\zeta$ is the base learning rate. It is noteworthy that all DLO components are trained during the Supervised Fine-Tuning (SFT) stage in an end-to-end manner, eliminating the need for Continual Pre-Training (CPT). By integrating DLO, we achieve dynamic vertical scaling, optimizing model depth, and maintaining high performance with reduced computational demands.

\subsection{Adaptive Inference-Time FLOPs}
During inference time, DLO uses layer-specific sparsity settings to maintain computational efficiency and ensure adaptive floating-point operations (FLOPs) for different tokens.
In other words, the predicted sparsity $\hat\rho_i$ will be determined completely by the router based on Equation~\eqref{eq:router}-\eqref{eq:output_calculation}.
The adaptive FLOPs are computed as:
\begin{equation}
    \text{FLOPs}_i = \hat\rho_i \cdot \text{FLOPs}_{\text{full}},
\end{equation}
where $\text{FLOPs}_{\text{full}}$ represents the FLOPs for a fully active layer. Since $\hat\rho_i$ is predicetd based on each specific token, DLO acheive adaptive FLOPs that entails better generalizability.

\section{Experiments}

In this section, we present an empirical evaluation of the proposed DLO framework, detailing the experimental settings, results, and analysis.

\subsection{Experimental Settings}
\label{sec:setting}

\textbf{Model Selection.}
We utilize LLaMA2-7B~\cite{touvron2023llama} as the primary backbone due to its open-source availability and extensive usage. It consists of $R = 32$ original transformer layers, which we group into $P=4$ clusters, each containing $Q=8$ layers. For layer expansion, we increase the group size to $Q'=10$ layers, resulting in a dense model, \llamaex, with a total of 40 layers and 8 billion parameters. For comparison, we also employ LLaMA-Pro-8B~\cite{wu2024llama}, a competitive model trained with Continual Pre-Training (CPT) on specialized datasets. We demonstrate DLO achieves an optimal balance between performance and computational cost in Section~\ref{sec:cost}.

\noindent\textbf{Fine-tuning Details.}
Following common practices \cite{wu2024llama}, we fine-tune using a mixture of five instruction tuning datasets: ShareGPT~\cite{Sharegpt}, EvolInstruct~\cite{luo2023wizardcoder}, SlimOrca~\cite{Slimorca}, MetaMath~\cite{yu2023metamath}, and Evol-CodeAlpaca~\cite{EvolCodeAlpaca}, with ShareGPT replicated three times, totaling approximately 1.44 million instances. We use a batch size of 128 and a maximum sequence length of 4,096 tokens. The learning rate is set to $2e-5$ with a warmup ratio of 0.03 and cosine scheduling, and we utilize AdamW~\cite{loshchilov2017decoupled} as the optimizer. Flash Attention~\cite{dao2205fast} and bfloat16 mixed-precision training are adopted to accelerate training. Fine-tuning \llamaex~under different skip ratios yields the sparse models, with each training run taking approximately 36 hours on eight NVIDIA A100 GPUs.

\noindent\textbf{Evaluation Benchmarks.}
We assess the fine-tuned models using the EleutherAI LM Harness~\cite{eval-harness} and BigCode Harness \cite{bigcode-evaluation-harness} across three domains: \ding{182} \textit{Language} [ARC-C~\cite{clark2018think}, GLUE~\cite{wang2018glue}, MMLU~\cite{hendrycks2020measuring}, OBQA~\cite{mihaylov2018can}, PIQA~\cite{bisk2020piqa}, SQuAD~\cite{rajpurkar2016squad}, TruthfulQA~\cite{lin2021truthfulqa}, WinoGrande~\cite{sakaguchi2021winogrande}], \ding{183} \textit{Math} [GSM8K~\cite{cobbe2021training}, MathQA~\cite{amini2019mathqa}], and \ding{184} \textit{Code} [HumanEval~\cite{chen2021evaluating}, MBPP~\cite{austin2021program}]. Detailed metrics are in Appendix~\ref{app:evaluation}.

\begin{table*}[t]
    \centering
    \resizebox{0.98\linewidth}{!}{
    \begin{tabular}{clc|cccccccc|cc|cc|c}
        \toprule
        \multirow{2}{*}{\textbf{Sparsity}} & \multirow{2}{*}{\textbf{Model}} & \multirow{2}{*}{\textbf{FLOPs}} & \multicolumn{8}{c|}{\textbf{Language}} & \multicolumn{2}{c|}{\textbf{Math}} & \multicolumn{2}{c|}{\textbf{Code}} & \multirow{2}{*}{\underline{\textbf{Avg.}}~$\uparrow$} \\
         & & & ARC-C & GLUE & MMLU & OBQA & PIQA & SQuAD & TruthfulQA & WinoGrande & GSM8K & MathQA & HumanEval & MBPP & \\
        \midrule
        \multirow{5}{*}{0\%}
         & \textcolor{orange}{$\bullet$} LLaMA2-7B & \multirow{2}{*}{29.3T}
         & 53.1 & 40.6 & 46.9 & 44.2 & 79.0 & \bf 26.4 & 38.8 & 74.0
         & 14.5 & 28.3 & 21.8 & 29.0 & \underline{41.38} \\
         & \textcolor{orange}{$\bullet$} LLaMA2-7B$_{+\text{\tiny SFT}}$ & 
         & 54.0 & 72.4 & 53.0 & 44.4 & 78.8 & 22.9 & \bf 40.8 & \bf 74.2
         & 56.6 & 30.8 & 57.3 & 30.5 & \bf \underline{51.31} \\
        \cdashline{2-16}
         & \textcolor{orange}{$\bullet$} LLaMA-Pro-8B & \multirow{2}{*}{36.4T}
         & \bf 54.1 & 40.7 & 47.9 & 41.6 & 78.2 & 14.2 & 39.0 & 74.0
         & 17.9 & 29.5 & 28.7 & \textbf{33.2} & \underline{41.58} \\
         & \textcolor{orange}{$\bullet$} LLaMA-Pro-8B$_{+\text{\tiny SFT}}$ & 
         & 51.0 & 71.0 & 53.0 & \bf 45.0 & 79.0 & 15.1 & 38.0 & 73.6
         & \bf 58.6 & 30.8 & \bf 58.4 & 30.5 &  \underline{50.33} \\
         \cdashline{2-16}
         & \textcolor{orange}{$\bullet$} \llamaex-8B$_{+\text{\tiny SFT}}$ & 36.5T
         & 53.2 & \bf 75.5 & \bf 53.2 & 43.7 & \bf 79.0 & 22.0 & 38.7 & 74.0
         & 57.4 & \bf 31.0 & 57.0 & 30.2 & \bf \underline{51.24} \\
        \midrule
        \multirow{2}{*}{10\%}
         & \textcolor{orange}{$\circ$} LLaMA2-7B & 27.5T
         & 51.0 & 72.5 & 51.1 & 40.2 & 78.0 & 21.0 & 39.3 & 71.0
         & 53.4 & 29.6 & 49.7 & 28.3 & \underline{48.76} \\
         & \textcolor{orange}{$\circ$} \llamaex-8B & 34.2T
         & \bf 52.5 & \bf 75.4 & \bf 51.4 & \bf 43.6 & \bf 78.7 & \bf 21.7 & \bf 41.0 & \bf 73.4
         & \bf 55.0 & \bf 31.4 & \bf 55.3 & \bf 29.1 & \bf \underline{50.71} \\
        \midrule
        \multirow{2}{*}{20\%}
         & \textcolor{orange}{$\circ$} LLaMA2-7B & 25.8T
         & 33.0 & 69.9 & 50.4 & 35.0 & 65.6 & 16.4 & 36.9 & 54.2
         & 1.0 & 24.5 & 0.0 & 0.0 & \underline{32.24} \\
         & \textcolor{orange}{$\circ$} \llamaex-8B & 32.0T
         & \bf 51.2 & \bf 73.3 & \bf 50.8 & \bf 43.2 & \bf 78.2 & \bf 20.4 & \bf 38.9 & \bf 73.2
         & \bf 50.1 & \bf 30.5 & \bf 57.6 & \bf 28.2 & \bf \underline{49.63} \\
        \midrule
        \multirow{2}{*}{30\%}
         & \textcolor{orange}{$\circ$} LLaMA2-7B & 24.1T
         & 28.1 & 2.8 & 47.1 & 35.0 & 53.8 & 13.9 & 37.9 & 52.2
         & 0.0 & 21.6 & 0.0 & 0.0 & \underline{24.40} \\
         & \textcolor{orange}{$\circ$} \llamaex-8B & 29.8T
         & \bf 44.6 & \bf 73.0 & \bf 50.1 & \bf 41.2 & \bf 77.1 & \bf 21.7 & \bf 37.1 & \bf 63.2
         & \bf 46.9 & \bf 28.1 & \bf 31.2 & \bf 6.0 & \bf \underline{43.35} \\
        \bottomrule
    \end{tabular}
    }
    \caption{Performance comparison of DLO (our approach) on various datasets using LLaMA2-7B as the backbone. Models marked with \textcolor{orange}{$\bullet$} are dense models, either original or those expanded using DLO \texttt{expansion}. Models with 8B parameters indicate expansion via LLaMA-Pro or our DLO. Models marked with \textcolor{orange}{$\circ$} are sparse models incorporating DLO \texttt{activation} and \texttt{skipping} operations. Inference FLOPs are counted with a sequence with 2,048 tokens. The proposed \textcolor{orange}{$\bullet$} \llamaex-8B~with $0\%$ spasity signifies that no layer is skipped and all the original and expanded layers are activated. \textcolor{orange}{$\circ$} LLaMA2-7B with non-zero sparsity equals \llamaex~without expanding layers. }
    \label{tab:compare}
    \vspace{-4mm}
\end{table*}

\vspace{-1mm}
\subsection{Overall Performance}\label{sec:main_result}

Table~\ref{tab:compare} summarizes the performance of the DLO framework across various datasets using LLaMA2-7B as the backbone. From the results, we draw several key observations are as follows:

\noindent\ding{182}~{\textbf{Dense Models' Superiority:}}
Dense models, indicated by \textcolor{orange}{$\bullet$}, generally outperform their sparse counterparts across most datasets. For instance, \llamaex~models consistently achieve high average scores, such as 51.31 for LLaMA2-7B+\textsubscript{SFT}, compared to the baseline LLaMA2-7B's 41.38. This indicates that our DLO-\texttt{expansion} method enhances model performance significantly while leveraging additional parameters effectively.
\noindent\ding{183}~{\textbf{Efficiency of Sparse Models:}}
Sparse models, marked with \textcolor{orange}{$\circ$}, show a notable reduction in inference-time FLOPs while maintaining competitive accuracy. At 10\% sparsity, the \llamaex~model achieves an average score of 50.71 with 34.2T FLOPs, compared to the dense model's 51.31 with 36.5T FLOPs. This demonstrates the efficiency of DLO's \texttt{activation} and \texttt{skipping} operations in optimizing computational resources without significantly sacrificing performance.
\noindent\ding{184}~{\textbf{DLO-\texttt{Expansion} Advantages:}}
Models expanded using DLO \texttt{expansion} with up to 8B parameters outperform the original LLaMA2-7B model across multiple metrics. For example, dense LLaMA-DLO-8B+\textsubscript{SFT} achieves a higher average score of 51.24 compared to LLaMA2-7B’s 41.38, highlighting the effectiveness of vertical scaling through layer expansion in improving model capacity and performance. On the other hand
\noindent\ding{185}~{\textbf{Balanced Performance of Sparse Models:}}
Sparse models with DLO's dynamic \texttt{activation} and \texttt{skipping} (\textcolor{orange}{$\circ$}) provide a well-balanced trade-off between performance and computational efficiency. At 30\% sparsity, \llamaex~models maintain strong performance on tasks like GLUE (51.0 vs. 28.1) and HumanEval (50.5 vs. 21.8), while significantly reducing FLOPs. This makes them suitable for scenarios requiring computational efficiency without substantial performance loss.
\noindent\ding{186}~{\textbf{Effective Inference Optimization:}}
DLO demonstrates effective inference optimization. For instance, the dense \llamaex~model with 8B parameters achieves lower inference-time FLOPs compared to LLaMA-Pro-8B (36.5T vs. 36.4T) while maintaining a competitive average performance (51.24 vs. 50.33). This highlights DLO's capability to enhance model efficiency without compromising accuracy.
\noindent\ding{187}~{\textbf{General Observations:}}
Overall, the DLO framework successfully balances performance and efficiency across various datasets and tasks. The adoption of both \texttt{expansion} and \texttt{skipping} strategies enables \llamaex~to achieve robust performance improvements while maintaining lower computational costs, suggesting that DLO is a viable approach for scalable and efficient LLM deployment. 

We conduct further analyses of key components of DLO in the subsequent subsections.

\subsection{Ablation Studies}
\label{sec:ablation}

\begin{table*}[t]
    \centering
    \resizebox{0.98\linewidth}{!}{
    \begin{tabular}{l|cccccccc|cc|cc|c}
        \toprule
        \multirow{2}{*}{\textbf{Method}} & \multicolumn{8}{c|}{\textbf{Language}} & \multicolumn{2}{c|}{\textbf{Math}} & \multicolumn{2}{c|}{\textbf{Code}} & \multirow{2}{*}{\underline{\textbf{Avg.}}~$\uparrow$} \\
         & ARC-C & GLUE & MMLU & OBQA & PIQA & SQuAD & TruthfulQA & WinoGrande & GSM8K & MathQA & HumanEval & MBPP & \\
        \midrule
         Random
         & 25.1 & 41.4 & 24.4 & 26.8 & 50.3 & \textbf{42.2} & 37.7 & 49.3
         & 0.3 & 20.3 & 0.0 & 1.3 & \underline{26.6} \\
         \cellcolor[HTML]{FFE2C3}Identity
         & \cellcolor[HTML]{FFE2C3}\textbf{52.5} & \cellcolor[HTML]{FFE2C3}\textbf{75.4} & \cellcolor[HTML]{FFE2C3}51.4 & \cellcolor[HTML]{FFE2C3}\textbf{43.6} & \cellcolor[HTML]{FFE2C3}\textbf{78.7} & \cellcolor[HTML]{FFE2C3}21.7 & \cellcolor[HTML]{FFE2C3}\textbf{41.0} & \cellcolor[HTML]{FFE2C3}\textbf{73.4}
         & \cellcolor[HTML]{FFE2C3}\textbf{55.0} & \cellcolor[HTML]{FFE2C3}\textbf{31.4} & \cellcolor[HTML]{FFE2C3}\textbf{55.3} & \cellcolor[HTML]{FFE2C3}\textbf{29.1} & \cellcolor[HTML]{FFE2C3}\underline{\textbf{50.7}} \\
         Copy
         & 52.0 & 71.3 & 52.3 & 43.2 & 78.6 & 23.3 & 40.8 & 73.0
         & 52.2 & 29.0 & 41.1 & 26.7 & \underline{48.6} \\
         Linear
         & 48.6 & 70.7 & \textbf{53.4} & 39.4 & 72.9 & 18.7 & 38.7 & 57.5
         & 29.7 & 25.7 & 24.0 & 0.7 & \underline{40.0} \\
         Slerp
         & 42.2 & 71.6 & 49.6 & 39.8 & 76.4 & 22.3 & 36.2 & 63.0
         & 43.0 & 27.1 & 40.4 & 4.9 & \underline{43.0} \\
        \bottomrule
    \end{tabular}
    }
    \caption{Experiments on the effectiveness of different initialization strategies for the expaned blocks. For this study we evaluate on \textcolor{orange}{$\circ$} \llamaex-8B with 10\% sparsity.}
    \label{tab:abl_expand_init}
    \vspace{-2mm}
\end{table*}

\begin{table*}
    \centering
    \resizebox{0.98\linewidth}{!}{
    \begin{tabular}{l|cccccccc|cc|cc|c}
        \toprule
        \multirow{2}{*}{\textbf{Method}} & \multicolumn{8}{c|}{\textbf{Language}} & \multicolumn{2}{c|}{\textbf{Math}} & \multicolumn{2}{c|}{\textbf{Code}} & \multirow{2}{*}{\underline{\textbf{Avg.}}~$\uparrow$} \\
         & ARC-C & GLUE & MMLU & OBQA & PIQA & SQuAD & TruthfulQA & WinoGrande & GSM8K & MathQA & HumanEval & MBPP & \\
        \midrule
         \cellcolor[HTML]{FFE2C3}DLO
         & \cellcolor[HTML]{FFE2C3}\textbf{52.5} & \cellcolor[HTML]{FFE2C3}\textbf{75.4} & \cellcolor[HTML]{FFE2C3}51.4 & \cellcolor[HTML]{FFE2C3}43.6 & \cellcolor[HTML]{FFE2C3}\textbf{78.7} & \cellcolor[HTML]{FFE2C3}\textbf{21.7} & \cellcolor[HTML]{FFE2C3}\textbf{41.0} & \cellcolor[HTML]{FFE2C3}73.4
         & \cellcolor[HTML]{FFE2C3}\textbf{55.0} & \cellcolor[HTML]{FFE2C3}\textbf{31.4} & \cellcolor[HTML]{FFE2C3}55.3 & \cellcolor[HTML]{FFE2C3}\textbf{29.1} & \cellcolor[HTML]{FFE2C3}\underline{\textbf{50.7}} \\
         w/o Zero Init
         & 51.9 & 73.0 & \textbf{52.1} & \textbf{44.6} & 77.7 & 20.8 & 40.3 & \textbf{74.3}
         & \textbf{55.0} & 30.5 & \textbf{56.1} & 28.2 & \underline{50.4} \\
         w/o Rescaling
         & 47.4 & 71.3 & 49.4 & 35.8 & 75.3 & 21.1 & 39.6 & 67.5
         & 35.0 & 24.5 & 55.6 & 28.2 & \underline{45.9} \\
        \bottomrule
    \end{tabular}
    }
    \caption{Ablation Study on the effectiveness of zeros router initialization \& score rescaling. For this evaluation, we deploy \textcolor{orange}{$\circ$} \llamaex-8B with 10\% sparsity for experiment.}
    \label{tab:abl_router}
    \vspace{-4mm}
\end{table*}

\begin{table}[htpb]
    \centering
    \resizebox{0.98\linewidth}{!}{
    \begin{tabular}{lcccccc}
        \toprule
         \textbf{Method} & ARC-C & MMLU & TruthfulQA & WinoGrande & GSM8K & \underline{Avg.}~$\uparrow$\\
        \midrule
         SOLAR
         & 24.8 & 24.8 & 38.8 & 50.7 & 2.2 & \underline{28.3} \\
         SD-Stack
         & 23.5 & 23.4 & 36.0 & 51.1 & 2.5 & \underline{27.3} \\
        \hdashline
         \cellcolor[HTML]{FFE2C3}DLO
         & \cellcolor[HTML]{FFE2C3}\textbf{52.5} & \cellcolor[HTML]{FFE2C3}\textbf{51.4} & \cellcolor[HTML]{FFE2C3}\textbf{41.0} & \cellcolor[HTML]{FFE2C3}\textbf{73.4} & \cellcolor[HTML]{FFE2C3}\textbf{55.0} & \cellcolor[HTML]{FFE2C3}\underline{\textbf{54.7}} \\
        \bottomrule
    \end{tabular}
    }
    \caption{Comparison with different expansion methods. We extend \textcolor{orange}{$\circ$}~\llamaex~layers using different strategies and fine-tune the expanded models with DLO under overall skip rate $\rho=10\%$. }
    \label{tab:abl_expand}
    \vspace{-5mm}
\end{table}

\paragraph{Initialization Strategies for Expanded Layers.}

We explored the effectiveness of various layer initialization strategies for the expanded layers, as detailed in Section~\ref{sec:expand}. Table~\ref{tab:abl_expand_init} evaluates the impact of different initialization strategies on the performance of \llamaex~models with 10\% sparsity. The results highlight that the choice of initialization plays a critical role in determining model performance across various tasks.

\noindent\ding{182}~The identity and copy initialization strategies demonstrate the most consistent and high-performing results, suggesting that leveraging existing layer information is beneficial for stabilizing and enhancing model performance. These methods help maintain coherence in the model’s internal representations, leading to robust results across a wide range of tasks, including GLUE and HumanEval.

\noindent\ding{183}~Interestingly, while linear and SLERP initializations were expected to offer smoother transitions and potentially enhance performance, their results were only moderately effective. This indicates that while sophisticated initialization techniques can offer benefits, they may not always outperform simpler strategies like identity and copy initialization, which directly utilize pre-existing model structures.

\noindent\ding{184}~Random initialization yields the lowest performance. The variability in task performance with this method highlights the challenges of using non-specific weights, which can lead to unstable and suboptimal model behavior, particularly in complex tasks like math and coding.

Overall, the findings emphasize that initialization strategies that leverage prior information from existing layers tend to provide a better foundation for training expanded models, leading to improved performance. We thus choose $\Pi_{identity}$ as the default initialization strategy.

\vspace{-2mm}
\paragraph{Zeros Router Initialization \& Score Rescaling.}
In this experiment, we investigate the impact of zeros router initialization and score rescaling on mitigating performance degradation.

\noindent$\rhd$~\textit{Zeros Router Initialization.} Initializing the router parameters to zero aims to start the model from a neutral state, avoiding any initial bias towards layer activation or skipping. This method allows the model to learn activation patterns from scratch without being influenced by predefined weights. Results in Table~\ref{tab:abl_router} indicate that this approach helps maintain balanced training dynamics and mitigates premature convergence, as reflected in the performance stability observed across tasks.

\noindent$\rhd$~\textit{Score Rescaling.} Score rescaling adjusts the routing scores to maintain them within a specific range, typically 0 to 1. This adjustment is intended to preserve gradient flow and prevent extreme activations, ensuring that the model remains responsive to training signals. Our findings suggest that score rescaling helps avoid over-activation of layers, leading to more efficient use of the model’s capacity.

The combined use of zeros router initialization and score rescaling appears to prevent performance degradation effectively. As shown in Table~\ref{tab:abl_router}, models with these techniques generally achieve more consistent accuracy and efficiency across various tasks. These results suggest that careful initialization and rescaling strategies are beneficial for maintaining robust performance during adaptation.

\begin{table*}[t]
    \centering
    \resizebox{0.98\linewidth}{!}{
    \begin{tabular}{l|cccccccc|cc|cc|c}
        \toprule
        \multirow{2}{*}{\textbf{Method}} & \multicolumn{8}{c|}{\textbf{Language}} & \multicolumn{2}{c|}{\textbf{Math}} & \multicolumn{2}{c|}{\textbf{Code}} & \multirow{2}{*}{\underline{\textbf{Avg.~$\uparrow$}}} \\
         & ARC-C & GLUE & MMLU & OBQA & PIQA & SQuAD & TruthfulQA & WinoGrande & GSM8K & MathQA & HumanEval & MBPP & \\
        \midrule
         Uniform
         & 36.0 & 61.7 & 48.7 & 36.6 & 58.4 & 47.9 & 36.8 & 68.2
         & 25.4 & 21.7 & 23.7 & 2.2 & \underline{38.9} \\
         \cellcolor[HTML]{FFE2C3}Layer-Wise
         & \cellcolor[HTML]{FFE2C3}\textbf{52.5} & \cellcolor[HTML]{FFE2C3}\textbf{75.4} & \cellcolor[HTML]{FFE2C3}\textbf{51.4} & \cellcolor[HTML]{FFE2C3}\textbf{43.6} & \cellcolor[HTML]{FFE2C3}\textbf{78.7} & \cellcolor[HTML]{FFE2C3}\textbf{21.7} & \cellcolor[HTML]{FFE2C3}\textbf{41.0} & \cellcolor[HTML]{FFE2C3}\textbf{73.4}
         & \cellcolor[HTML]{FFE2C3}\textbf{55.0} & \cellcolor[HTML]{FFE2C3}\textbf{31.4} & \cellcolor[HTML]{FFE2C3}\textbf{55.3} & \cellcolor[HTML]{FFE2C3}\textbf{29.1} & \cellcolor[HTML]{FFE2C3}\underline{\textbf{50.7}} \\
        \bottomrule
    \end{tabular}
    }
    \caption{\small Performance of the fine-tuned \textcolor{orange}{$\circ$} \llamaex-8B with $10\%$ sparsiy and different sparsity distribution strategies. ``Uniform'' represents all layers use the same sparsity $\rho_i=\rho$ during training. ``Layer-Wise'' denotes the model maintains different skip rates for different layers, as described in Section \ref{sec:dlo_training}. 
    }
    \label{tab:abl_skip_ratee}
    \vspace{-1mm}
\end{table*}

\begin{figure*}[t]
    \centering
    \scalebox{1}{
\includegraphics[width=\linewidth]{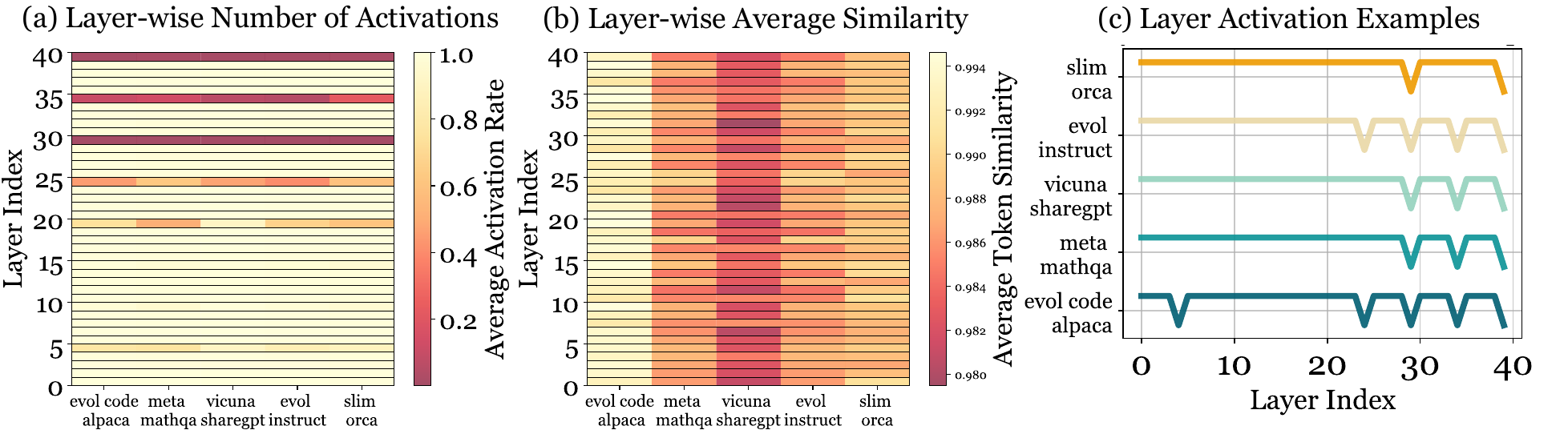}}
    \caption{\small Visualization on different datasets of (a) Layer-Wise Number of Activations, (b) Layer-Wise Average Similarity, and (c) Token Activation Examples.}
    \label{fig:skip_plots}
    \vspace{-4mm}
\end{figure*}

\begin{figure}[t]
    \centering
    \scalebox{1}{
\includegraphics[width=\linewidth]{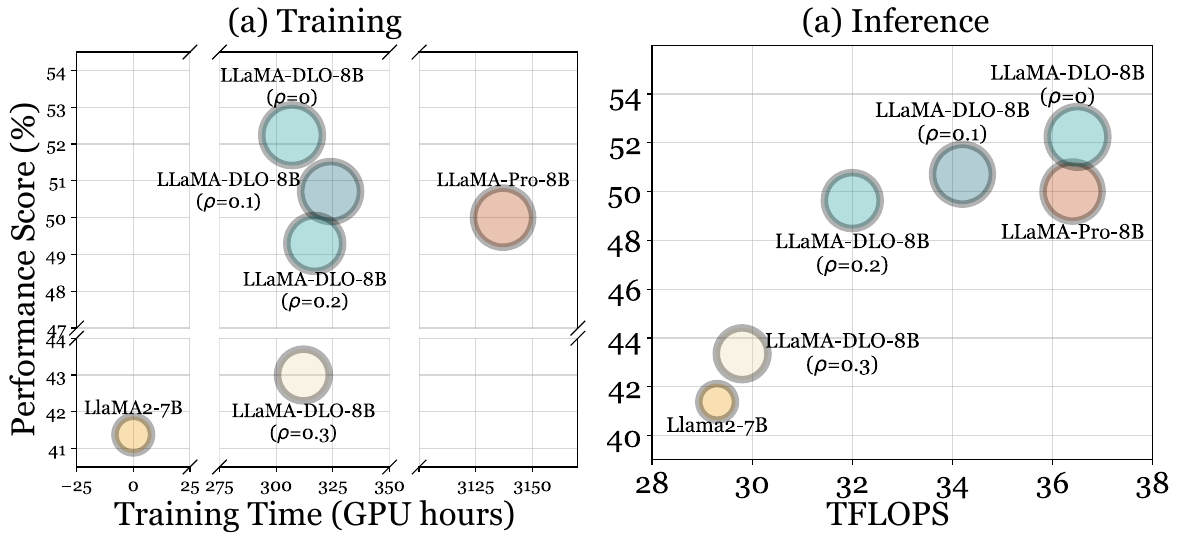}}
\vspace{-6mm}
    \caption{\small (a) Performance \textit{v.s.} Training
time. LLaMA-Pro is reported in H800 GPU hours quoted from the original paper. The rests are reported in A100 GPU hours. (b) Performance \textit{v.s.} Inference FLOPs. DLO achieves the best trade-off between performance and training or inference costs.}
    \label{fig:scater}
    \vspace{-5mm}
\end{figure}

\vspace{-2mm}
\paragraph{Efficient Expansion.} 
In addition to the high-cost LLaMA-Pro approach (studied in Table~\ref{tab:compare}), we compare our expansion method with two state-of-the-art efficient vertical expansion baselines: SOLAR~\cite{kim2023solar} and Self-Duplicate Stack (SD-Stack)~\cite{sdstack}. These two methods duplicate blocks of transformer layers and stack them together in a training-free manner. 
As shown in Table~\ref{tab:abl_expand}, the proposed DLO-\texttt{expansion} significantly outperforms both SOLAR and SD-Stack by a considerable margin. This highlights the critical role of Supervised Fine-Tuning (SFT) in adapting the expanded layers effectively. Unlike training-free approaches, DLO-\texttt{expansion} achieves a superior balance between training cost and performance, demonstrating the importance of fine-tuning in maximizing the  effectiveness of layer expansion.

\vspace{-1mm}
\paragraph{Layer-Wise Skip Rates \& Sparsity Allocation.}

This experiment evaluates the impact of using layer-specific skip rates on sparsity allocation.

\noindent$\rhd$~\textit{Layer-Wise Skip Rates.} Adjusting skip rates for each layer aims to selectively activate or skip layers based on their contribution to task performance, which is measured by the layer-wise similarity. This method helps focus computational resources on more critical layers. Results in Table~\ref{tab:abl_skip_ratee} suggest that this approach can lead to more efficient sparsity allocation with less impact on model performance. Figure~\ref{fig:skip_plots} also show that DLO can skip layers that have high layer-wise
similarity.

\noindent$\rhd$~\textit{Sparsity Allocation.} Tailoring skip rates by layer helps distribute sparsity more effectively, potentially reducing computational overhead. As indicated in Table~\ref{tab:abl_skip_ratee}, models with layer-wise skip rates tend to maintain performance while achieving better computational efficiency.




\vspace{-1mm}
\subsection{Scalability}\label{sec:cost}
\vspace{-1mm}
The proposed \llamaex~model surpasses the performance of the original dense LLaMA, while also achieving comparable results to the dense LLaMA-Pro. Notably, it does so at a significantly lower training cost by eliminating the need for expensive CPT. Additionally, \llamaex~facilitates efficient inference through adaptively reduced FLOPs, making it a cost-effective choice for both training and deployment.

Figure~\ref{fig:scater} illustrates the trade-off between model performance and both training and inference costs. \llamaex~emerges as the optimal solution, achieving the best balance across these metrics. This demonstrates the model's scalability, ensuring that high performance is maintained while keeping computational costs manageable.

\vspace{-2mm}
\section{Conclusion}
\vspace{-2mm}
This paper presents \llamaex, a framework for efficient vertical scaling of LLMs that dynamically expands, activates, and skips layers to optimize computational resources. Our experiments demonstrate that \llamaex~achieves performance on par with expensive dense expansion model like LLaMA-Pro, while significantly reducing training costs and enhancing inference efficiency. These results highlight \llamaex's potential as a cost-effective solution for scaling LLMs in various NLP tasks, offering a balanced approach between model performance and resource management.

\section*{Limitation Discussions \& Future Work}

\paragraph{Disentanglement of Routing Decisions and Rescaling Scores.}
Currently, the routing decisions and rescaling scores in our framework are interdependent, which may impact the model's accuracy. For instance, the skipped outputs are optimized to match the original outputs primarily through cosine similarity, which does not account for the difference in L2 magnitude. This discrepancy could potentially be mitigated by applying an additional rescaling factor to the skipped outputs, ensuring a better match in magnitude and improving the overall performance.

\paragraph{Improved Supervision for Router Labels.}
The current method relies on cosine similarity for supervising router labels, which may not be the most effective approach. Exploring alternative supervision methods, such as task-specific metrics or direct gradients, could lead to more accurate router decisions. For example, drawing inspiration from works like \citet{jiang2023llmlingua} or employing gradient-based techniques similar to those used in network pruning, could enhance the router's ability to prioritize important tokens and improve performance.

\paragraph{Skipping Attention Modules.}
This work primarily focuses on skipping MLP layers due to the observed instability in decoding when skipping attention modules, such as excessive repetition and degraded accuracy (e.g., achieving $0\%$ accuracy on GSM8K). Future work could explore strategies to stabilize the skipping of attention layers, potentially improving model efficiency without compromising output quality significantly.

\section*{Ethical Statement}

The development and deployment of large language models, including the \llamaex~framework presented in this work, can raise important ethical considerations. Our research aims to enhance the efficiency and scalability of LLMs while maintaining high standards of responsibility and ethical practice. We recognize the potential impact of our work on various stakeholders and are committed to the following ethical principles:

\paragraph{Fairness and Bias Mitigation.} We are aware that language models can inadvertently learn and propagate biases present in training data. Efforts have been made to ensure that \llamaex~is trained on diverse and representative datasets to minimize the risk of bias. Effective ways to mitigate bias is a pressing problem worth further study.

\paragraph{Transparency and Accountability.} We strive to maintain transparency in our research and development processes. Detailed documentation and open access to our methodologies and results will be provided to allow for scrutiny and reproducibility. Accountability mechanisms are in place to ensure that any adverse effects of our technology are promptly identified and addressed.

\paragraph{Social Impact.} The potential societal impact of \llamaex~is carefully evaluated to prevent misuse or harm. We are committed to the ethical deployment of our technology, ensuring it is used for beneficial purposes such as advancing research, improving accessibility, and enhancing communication. We actively discourage and take steps to prevent the use of our models for malicious activities, misinformation, or any application that could harm individuals or society.

\paragraph{Continuous Ethical Review.} The ethical implications of our work are continually assessed to adapt to evolving norms and expectations. We engage with interdisciplinary experts and stakeholders to identify and address ethical concerns, ensuring that our research and its applications remain aligned with societal values and ethical standards.

By adhering to these principles, we aim to contribute positively to the field of artificial intelligence and ensure that the benefits of our research are realized in an ethical and responsible manner.



\newpage
\bibliography{ref}
\bibstyle{acl_natbib}

\newpage
\appendix

\section{Pseudo Code style Description of Dynamic Layer Operation (DLO)}
\label{app:code}

\begin{algorithm}
\caption{Dynamic Layer Operation (DLO)}
\begin{algorithmic}[1]
\small
\REQUIRE Pre-trained LLM with $R$ layers, group size $Q$, expansion size $q$, target overall sparsity $\rho$, training steps $T$, annealing steps $T'$, base learning rate $\bar\zeta$
\ENSURE Optimized LLM with dynamic scaling
\STATE \textbf{Initialize:} $P \leftarrow R \ /\ Q$, $Q' \leftarrow Q + q$, $\rho_{i,1} \leftarrow \rho$
\STATE \textcolor{gray}{\text{// }\texttt{Layer Expansion:}}
\FOR{group $i \leftarrow 1$ to $P$}
    \FOR{layer $j \leftarrow Q+1$ to $Q+q$}
        \STATE Initialize $\theta'_{ij}$ using $\Pi$:
        \IF{$\Pi = $ `Xavier'}
            \STATE $\theta'_{ij} \sim \mathcal{U}\left(-\sqrt{\frac{6}{n_{in} + n_{out}}}, \sqrt{\frac{6}{n_{in} + n_{out}}}\right)$
        \ELSIF{$\Pi = $ `Copy'}
            \STATE $\theta'_{ij} \mathcal{L}_{\text{skip}} \theta_{i(Q+q-1)}$
        \ELSIF{$\Pi = $ `Identity'}
            \STATE $\theta'_{ij} \mathcal{L}_{\text{skip}} \theta_{i(Q+q-1)}$, $W'_{\text{out}} = 0$
        \ELSIF{$\Pi = $ `Linear Merge'}
            \STATE $\theta'_{ij} \mathcal{L}_{\text{skip}} \sum_{k=1}^{\tau} \alpha_k \theta_{i(Q+q-k)}$
        \ELSIF{$\Pi = $ `SLERP'}
            \STATE $\Omega = \arccos \left( \frac{\mathbf{u} \cdot \mathbf{v}}{\|\mathbf{u}\| \|\mathbf{v}\|} \right)$
            \STATE $\theta'_{ij} \mathcal{L}_{\text{skip}} \frac{\sin((1 - \alpha) \Omega)}{\sin(\Omega)} \mathbf{u} + \frac{\sin(\alpha \Omega)}{\sin(\Omega)} \mathbf{v}$
        \ENDIF
    \ENDFOR
\ENDFOR

\STATE \textcolor{gray}{\text{// }\texttt{Layer Activation and Skipping:}}
\FOR{step $t \leftarrow 1$ to $T$}
    \STATE \textcolor{gray}{\text{// }\texttt{Skip Rate Annealing:}}
    \IF{$t \leq T'$}
        \STATE $\rho^{t} \leftarrow \bar{\rho} + (\rho - \bar{\rho}) \frac{t}{T'}$
    \ELSE
        \STATE $\rho^{t} \leftarrow \rho$
    \ENDIF
    
    \STATE \textcolor{gray}{\text{// }\texttt{Training and Integration:}}
    \STATE $\mathcal{L}_{\text{skip}} \leftarrow 0$
    \FOR{layer $i \leftarrow 1$ to $R'$}
        \FOR{token $s$ in sequence}
            \STATE \textcolor{gray}{\text{// }\texttt{Dynamic Skip:}}
            \STATE $r_i^s \leftarrow \frac{1}{2} (\beta+(2\sigma(\mathbf{h}_i^s W_i) - 1)\gamma)$
            \IF{training}
                \STATE $\hat\lambda_i^s \leftarrow \mathbb{1}{ r_i^s \in \mathrm{Top}_{ \lfloor\rho_{i,t} S\rfloor } \big(\{r_i^s\}_{s=1}^S\big) }$
            \ELSE
                \STATE $\hat\lambda_i^s \leftarrow \mathbb{1}{ r_i > \frac{\beta}{2} }$
            \ENDIF
            \IF{$r_i^s = 1$}
                \STATE $\mathbf{h}_{i+1}^s \leftarrow r_i \cdot \mathcal{M}_i \circ \mathcal{A}_i(\mathbf{h}_i^s)$
            \ELSE
                \STATE $\mathbf{h}_{i+1}^s \leftarrow \mathcal{A}_i(\mathbf{h}_i^s)$
            \ENDIF
    
            \STATE \textcolor{gray}{\text{// }\texttt{Skip Loss:}}
            \STATE $\mu_i^s \leftarrow \cos(\mathcal{A}_i(\mathbf{h}_i^s), \mathcal{M}_i \circ \mathcal{A}_i(\mathbf{h}_i^s))$
            \STATE $\tilde\lambda_i^s \leftarrow \mathbb{1}{ \mu_{i}^s \in \mathrm{Bottom}_{\lfloor(1-\rho^t)R'S\rfloor}\big(\{\mu_{i}^s\}_{i,s=1}^{R',S}\big) }$
            \STATE $\mathcal{L}_{\text{skip}} \leftarrow \mathcal{L}_{\text{skip}} + \mathcal{L}_{\text{BCE}}(\sigma(\mathbf{h}_i^sW_i), \tilde{\lambda}_i^s)$
        \ENDFOR

        \STATE \textcolor{gray}{\text{// }\texttt{Layer-Wise Skip Rate:}}
        \STATE $\rho_{i,t+1} \leftarrow \sum_{s=1}^S \Tilde\lambda_i^s \ /\ S$
    \ENDFOR
    \STATE $\mathcal{L}_{\text{skip}} \leftarrow \mathcal{L}_{\text{skip}} \ /\ R'S$
    
    \STATE $\mathcal{L} \leftarrow \mathcal{L}_{\text{task}} + \mathcal{L}_{\text{skip}}$ 
    \STATE Adjust learning rate $\zeta_{i,t} \leftarrow \bar\zeta \cdot \frac{1 - \rho_{i,t}}{1 - \rho^{t}}$
\ENDFOR

\RETURN Optimized LLM
\end{algorithmic}
\end{algorithm}

\section{Evaluation Details}
\label{app:evaluation}
We list the number of shots and the metric used for each dataset as follows:

\begin{itemize}
    \item ARC-C: 25 shots, normalized accuracy.
    \item GLUE: 0 shot, accuracy.
    \item MMLU: 5 shots, normalized accuracy.
    \item PIQA: 0 shot, normalized accuracy.
    \item OBQA: 0 shot, normalized accuracy.
    \item SQuAD: 0 shot, F1 Score.
    \item TruthfulQA: 0 shot, accuracy.
    \item WinoGrande: 5 shots, accuracy.
    \item GSM8K: 5 shots, accuracy.
    \item MathQA: 0 shot, normalized accuracy.
    \item HumanEval: 200 rounds, pass@100.
    \item MBPP: 15 rounds, pass@10.
\end{itemize}

\section{Acknowledgment of AI Assistance in Writing and Revision}
We utilized ChatGPT-4 for revising and enhancing sections of this paper.

\end{document}